\documentclass[runningheads]{llncs}
\usepackage{graphicx}
\usepackage{amsfonts}
\usepackage{multirow}

\begin{document}
\title{Improving Interpretability for Computer-aided Diagnosis tools on Whole Slide Imaging with Multiple Instance Learning and Gradient-based Explanations}
\author{Antoine Pirovano\inst{1,2} \and
Hippolyte Heuberger\inst{1} \and
Sylvain Berlemont\inst{1} \and
Sa\"{i}d Ladjal\inst{2} \and
Isabelle Bloch\inst{2}}

\authorrunning{A. Pirovano et al.}
\institute{Keen Eye, France\\ 
\email{antoine.pirovano@keeneye.tech, hippolyte.heuberger@keeneye.tech, sylvain.berlemont@keeneye.tech}\\ \and
LTCI, T\'el\'ecom Paris, Institut Polytechnique de Paris, France\\
\email{said.ladjal@telecom-paristech.fr, isabelle.bloch@telecom-paris.fr}\\}
\maketitle              % typeset the header of the contribution
\begin{abstract}
Deep learning methods are widely used for medical applications to assist medical doctors in their daily routines.
While performances reach expert's level, interpretability (highlight how and what a trained model learned and why it makes a specific decision) is the next important challenge that deep learning methods need to answer to be fully integrated in the medical field.
In this paper, we address the question of interpretability in the context of whole slide images (WSI) classification.
We formalize the design of WSI classification architectures and propose a piece-wise interpretability approach, relying on gradient-based methods, feature visualization and multiple instance learning context. We aim at explaining how the decision is made based on tile level scoring, how these tile scores are decided and which features are used and relevant for the task.
After training two WSI classification architectures on Camelyon-16 WSI dataset, highlighting discriminative features learned, and validating our approach with pathologists, we propose a novel manner of computing interpretability slide-level heat-maps, based on the extracted features, that improves tile-level classification performances by more than $29\%$ for AUC.

% Contributions:
% Formalize approach of WSI pipelines
% Understanding of how these pipeline learn
% Understanding why ImageNet transfer works well on histopathological problems
% Quantification of decision (measure of interpretability) and of patches illustration relevancy
% Propose a priori identification of usefull features  
% Usefulness validated by pathologists

\end{abstract}
\section{Introduction}
\label{intro}
% Deep learning and computer aided medical decisions
Since their successful application for image classification \cite{Krizhevsky12} on ImageNet \cite{Deng09}, deep learning methods (especially Convolutional Neural Network (CNN) deep architectures) have been extensively used and adapted to tackle efficiently a wide range of health issues \cite{Pasa19,Pratt16}.

% WSIs 
Along with these new methods, the recent emergence of Whole Slide Imaging (WSI), microscopy slides digitized at a high resolution, represents a real opportunity for the development of efficient Computer-Aided Diagnosis (CAD) tools to assist pathologists in their work. Indeed, over the last three years, notably due to the WSI publicly available datasets, such as Camelyon-16 \cite{Ehteshami17} and TCGA~\cite{Tomczak15}, and in spite of the very large size of these images (generally around 10 giga pixels per slide), deep learning architectures for WSI classification have been developed and proved to be really efficient. 

% WSI Classification pipelines
In this work, we are interested in WSI classification architectures that use only the global label (e.g. diagnosis) to train and require no intermediate information such as cell labeling or tissue segmentation (which are time-consuming annotations). The training is regularized by introducing prior knowledge by design in the architectures which, in addition, makes the result interpretable. But the interpretability beyond the architectural design is still pretty shallow.

% Interpretability and explainability
However, interpretability (capacity to provide explanations that are relevant and interpretable by experts in the field) for medical applications are critical in many ways. (i) For routine tools where useful features are well known and are subject to a consensus among experts, it is important to show that the same features are used by the trained model in order to gain confidence of practitioners. (ii) A good explainability would enable to get the most out of the architectural interpretability and thus assist more efficiently medical doctors in their slide reviews. (iii) the ability to train using only slide level supervision opens a new field we call discovery which consists in predicting, based on easier access (e.g. less intrusive) data, outputs that generally requires heavy processes or waiting such as surgery (e.g. prognosis, treatment response). In order to be able to guide experts towards new discoveries the need for reliable interpretability is obviously high.

% Here
In this work, after formalizing the architectural design of most WSI architectures, we propose a piece-wise interpretability approach, that provides cell-level features that prove to be highly relevant and interpretable by pathologists. We also propose a new way of computing explanation slide-level heat-maps based on cell-level identified features and measure their interpretability relevance.
%%%%%%%%%%%

\section{Related Work and Motivations}
\label{relatedwork}

All successful WSI classification architectures deal with these very large images by cutting them into tiles, which is close to the workflow of pathologists who generally analyze these slides at levels of magnification between 5X and 40X.

% MIL
Recently, as explained in Section \ref{intro}, architectures that are able to learn using only global slide-level label have been proposed. They rely on a context of Multiple Instance Learning (MIL) i.e. slides are represented by bags of tiles with positive bags containing at least one positive tile and negative bags containing only negatives tiles. For example, CHOWDER \cite{Courtiol18} is an extension of WELDON \cite{Durand16} solution for WSI classification that uses min-max strategy to guide the training and make the decision. This approach reaches an AUC of 0.858 on Camelyon-16 and 0.915 on TCGA-lung (subset of TCGA dataset related to lung cancer). In~\cite{Ilse18}, an attention module \cite{Raffel15} is used instead of a min-max layer. AUC of 0.775 for a breast cancer dataset and 0.968 for a colon cancer dataset were reported. Recently, more works on large datasets proposed architectures that follow the same design \cite{Campanella19,Li19}. Heat-maps based on intermediate scores computed in these architectures are what we call architectural explainability that results from prior knowledge on WSI problems that is introduced by design in the architecture. They are of great interest and have proved to be really efficient to the point of being able to spot cancerous lesions that had been missed by experts (in \cite{Campanella19}). However explanations are relying on a single ``medical" score which might limit the interpretability regarding complex tissue structures that can be found on these slides.

While interpretability for deep learning CNN models is still at its beginning, some methods arise from the literature. ``Feature Visualization" has been extensively developed in \cite{Olah18}. It consists of a group of methods that aims at outputting visualizations to express in the most interpretable manner features associated with a single neuron or a group of neurons. It can be used to understand the general training of a model. For example, the question of transferring features learned from natural images (ImageNet) to medical images has only recently been investigated \cite{Raghu19} while widely used and yet surprisingly good. It has also been used  to measure how robust a learned feature is \cite{Couteaux19}. Other explainability methods are called attribution methods i.e. methods that output values reflecting, for each input, its contribution to the prediction. They are performed either through perturbation \cite{Fong17} or gradient computation (i.e. measure of the gradient of the output with respect to the input). This second group of methods is gaining more and more interest. In \cite{Simoyan13}, the authors show that gradient is a good approximation of the saliency of a model and even put forward a potential to perform weakly supervised localization. This work opened a new way of accessing explanations in deep neural networks and motivated a lot of interesting researches~\cite{Sundararajan17,Smilkov17,Goh20}. Mixed together these explanation methods can provide meaningful and complementary interpretability. 

To the best of our knowledge, a lot of explainability is still to be introduced in WSI classification architectures. In the next section, we present our approach to improve interpretability of a model trained for WSI classification in histopathology. We rely on gradient-based methods to identify and attribute the importance of features in intermediate descriptors, and on patch visualization for cell-level feature explanations. We also extend feature explanation to a slide level, thus drastically improving tumor localization and medical insights.

%%%%%%%%%%%%%%%%%%%%%%%%%%%%%%%%%%%%%%%%%%%%%%%%%%%

\section{Proposed Methods}
\label{contrib}

As introduced in Section \ref{relatedwork}, WSI classification architectures have a common design that we formalize here. 
Let $i$ be the slide index and $j$ the tile index for each slide. 
There are four distinct blocks in a typical WSI classification architecture:
\begin{enumerate}
\item A feature extractor module $f_{e}$ (typically a CNN architecture) that encodes each tile $x_{i,j}$ into a descriptor $d_{i,j} \in \mathbb{R}^N$ with $N$ the descriptor size (depending on the feature extractor):
$d_{i,j} = f_{e}(x_{i,j})$;

\item A tile scoring module $f_{s}$ that, based on each tile descriptor $d_{i,j}$, assigns a single score per tile $s_{i,j} \in \mathbb{R}$:
$s_{i,j} = f_{s}(d_{i,j})$;

\item An aggregation module $f_{a}$ that, based on all tile scores $s_{i,j}$, and sometimes their tile descriptors $d_{i,j}$, computes a slide descriptor $D_i \in \mathbb{R}^M$ with $M$ the slide descriptor size (depending on the aggregation module):
$D_i = f_{a}(s_{i,j},d_{i,j})$;

\item A decision module $f_{cls}$ that, based on the slide descriptor $D_i$, makes a class prediction $P_i \in \mathbb{R}^C$ with $C$ the number of classes:
$P_i = f_{cls}(D_i)$.
\end{enumerate}

\begin{figure}[!th]
\centering
\includegraphics[width=0.8\textwidth]{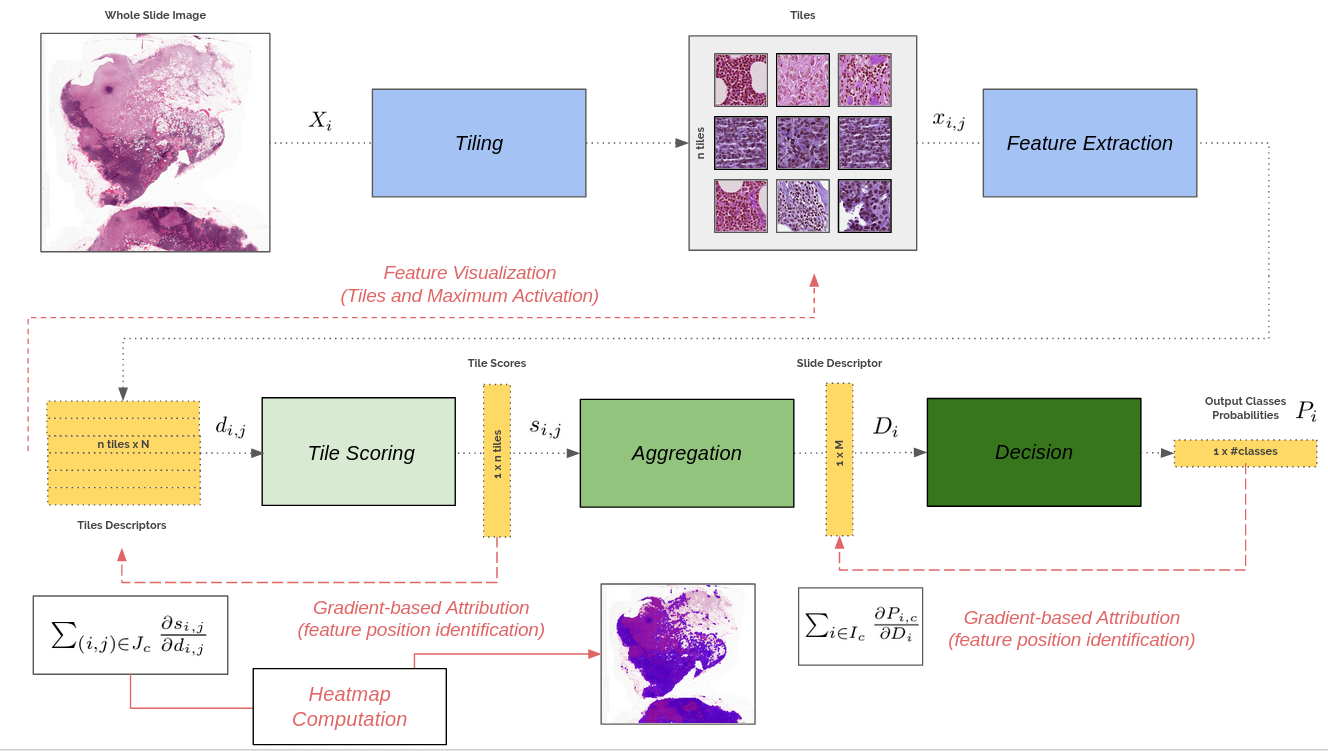}
\caption{Overview of the proposed method.}
\label{proposed_method}
\end{figure}

% Questions where to dig and how to dig (piece-wise interpretability "backpropagation")
Our approach (illustrated in Figure \ref{proposed_method}) consists in rewinding explanations from the decision module to tile information by applying interpretability methods and by answering successively the following three questions:
\begin{enumerate}
\item Which features of slide descriptors are relevant for a class prediction?
\item With regards to the aggregation module, which features of tile descriptors are responsible for previously identified relevant slide descriptor features?
\item Are these features of tile descriptors relevant medically and representative of histopathological information?
\end{enumerate}

The first question is answered using attribution vectors $A_{c} \in \mathbb{R}^M$ (one for each class $c$) computed as the gradient of the component of index $c$ of $P_i$ (noted $P_{i,c}$) with respect to $D_i$. It enables us to identify a set of relevant positions $K_c = \{K_{c,1}, ..., K_{c,L}\}$ in slide descriptors i.e. the $L$ (empirically determined) positions in $A_{c}$ with highest attributions over the slide predicted in class $c$: 

$A_{c} = \sum_{i \in I_c} \mid \frac{\partial P_{i,c}}{\partial D_i} \mid = \sum_{i \in I_c} \mid \frac{\partial f_{cls}(D_i)_c}{\partial D_i} \mid$,

\noindent with $I_c$ the set of slides predicted to be in class $c$.

Then, the second question is also answered using an attribution vector $a_{c} \in \mathbb{R}^N$ computed as the gradient of tile score $s_{i,j}$ with respect to tile descriptor $d_{i,j}$. This enables to identify features positions $k_c = \{k_{c,1}, ..., k_{c,l}\}$ in tile descriptors i.e. the $l$ (empirically determined) tile descriptors that are responsible for high activation at previously identified $K_c$ positions in slide descriptor:

$a_{c} = \sum_{(i,j) \in J_c} \frac{\mid \partial s_{i,j}}{\partial d_{i,j}} \mid = \sum_{(i,j) \in J_c} \mid \frac{\partial f_{s}(d_{i,j})}{\partial d_{i,j}} \mid$

\noindent with $J_c$ the set of tile positions $(i,j)$ that most activate $K_c$ positions in slide descriptors.

To answer the third question, we rely on feature activation to highlight features identified as being discriminative to the task by selecting tiles $x_{i,j}$ that have the highest activation per feature in $k_{c}$ identified over the whole test set. Along with these tiles, we display a maximum activation $\mathcal{X}$ image obtained by iteratively tuning pixels values to activate the feature by gradient ascent as follows: for each k in $k_{c}$, $\mathcal{X}$ is initialized as a uniformly distributed image $\mathcal{X}_0$; then while $f_{e}(\mathcal{X}_{n-1})_{k}$ increases, iterate over $n>0$: 

$\mathcal{X}_{n}(k) = \mathcal{X}_{n-1} + \frac{\partial f_{e}(\mathcal{X}_{n-1})_{k}}{\partial \mathcal{X}_{n-1}}$.

Finally, we also propose a new way to compute heat-maps for each slide $i$. We note $H_{c,i}$ the map that highlight regions on slide $i$ that explain what has been learned to describe class $c$ based on the identified features. For each slide $i$  and tile $j$, the heat-map value $H_{c,i,j}$ is computed as the average of activations $d_{i,j,k}$ (normalized per feature over all tiles of all slides) over identified features $k$ in $k_c$ for class $c$:

$H_{c,i,j} = \frac{1}{|k_{c}|} . \sum_{k \in k_{c}} \frac{d_{i,j,k}-\min_k}{\max_k-\min_k}$

\noindent with $\max_k = \max_{i,j}(d_{i,j,k})$ and $\min_k = \min_{i,j}(d_{i,j,k})$.

This heat-map (between 0 and 1) can be considered as a prediction scoring system and enables us to compute Area Under the ROC (Receiver Operating Characteristic) Curve to measure how relevant is the interpretability brought by our automatic feature extraction approach using ground truth lesion annotations when given.

%%%%%%%%%%%%%%%%%%%%%%%%%%%%%%%%%%%%%%%%%%%%%%%%

\section{Experiments and Results}

{\bf Architectures.}
% CHOWDER
% Attention-based
We validate our approach on two WSI classification trained architectures: CHOWDER and Attention-based classification. 

CHOWDER \cite{Courtiol18} uses a $1 \times 1$ convolution layer to turn each tile descriptor into a single tile score that are then aggregated using a min-max layer, that keeps the top-R and bottom-R scores (e.g. empirically $R$ = 5 gives the best results), to give a slide descriptor ($M = 2 \times R$).

Attention-based architecture \cite{Ilse18} uses an attention module (two $1 \times 1$ convolution layers with respectively 128 and 1 channels and a softmax layer) to compute competitive and normalized (sum to 1) tile scores from tile descriptors. Then, the slide descriptor is computed as the weighted (by tile scores) sum of tile descriptors ($M = N$).

Note that in our experiments the feature extractor is a ResNet-50 \cite{He15} ($N$ = 2048) trained on ImageNet and the decision module is a two layers fully connected network with 200 and 100 hidden neurons, respectively.

% Camelyon-16
\noindent {\bf Datasets.}
\label{datasets}
We validate our approach using Camelyon-16 dataset that contains 345 WSI divided into 209 ``normal" cases and 136 ``tumor" cases. This dataset contains slides digitized at 40X magnification from which we extract, with regard to a non-overlapping grid, $224 \times 224$ pixels at 20X magnification and pre-compute 2048-tile descriptors for each tile (using the ResNet-50 model trained on ImageNet). 216 slides are used to train our models while 129 slides form the test set to evaluate performances of models. 

\noindent {\bf Results on CHOWDER.}
Both architectures trained on Camelyon-16 show similar classification performances (AUC of 0.82 for the CHOWDER model and 0.83 for the Attention-based model). Let us now illustrate and detail the results of our approach on the CHOWDER model guided by the three questions raised in Section \ref{contrib}.

% Gradient-based explanation for Decision/Classification Module
The first question is ``Which slide descriptors features are relevant for a class prediction?" i.e. for CHOWDER given the $M$=10 ($R$=5) tile scores given as slide descriptor (the 5 minimum tile scores and the 5 maximum tile scores), what is the contribution of each of these values to the prediction?

The distribution of the (5-)min and (5-)max scores w.r.t. predictions over the whole 129 test slides shows that min scores are the ones that contribute to discriminate between the two classes (i.e. the lower min scores, the more the slide is predicted as being ``tumor"). A Mann-Whitney U-Test between scores (min and max independently) distributions reveals that min scores distributions per predicted class are statistically different ($p < 10^{-3}$) while max scores are not ($p = 0.23$). The attribution of min and max scores distributions validates this assertion.

% Tile Scoring Module
After explaining that min scores are the ones describing tumorous regions and thus that max scores are used for the ``normal" class, we are interested in identifying which features of tile descriptors are mostly responsible for minimum and maximum scores i.e. to describe each class. To address this second question, we use the same gradient-based explanation method.

Most minimal tile scores are under -5 and most maximal tile scores are above 11. For each of these groups of tiles, we compute the average attribution of each of the $N$=2048 features in tile descriptors (extracted by a ResNet-50 trained on ImageNet). The distribution of features hence activated allows us to identify which features are mostly responsible for min and max tile scores, i.e. highest attribution for min and max scored tiles. 

Thus we are able to claim that features (defined by their position in the descriptor) that are mostly useful for the trained model for each class are: 242, 420, 602, 1154, 1644, 1652 and 1866 ``tumor" class, and 565, 628, 647, 1158 and 1247 for ``normal" class. 

\noindent {\bf Interpretability.}
As exposed in the previous paragraph, based on explanations on decision blocks, we have been able to identify 7 and 5 features that are mostly used by the trained CHOWDER model to make decisions (and we did the same for the attention-based model). Now, we are interested in interpretable information to return to pathologists so that they can use their expertise to understand what these features put forward histopathologically speaking. We benefited from discussions with two experienced pathologists and report their overall feedback on the interpretable visualization we proposed.

\begin{figure}[!th]
\centering
\includegraphics[width=0.8\textwidth]{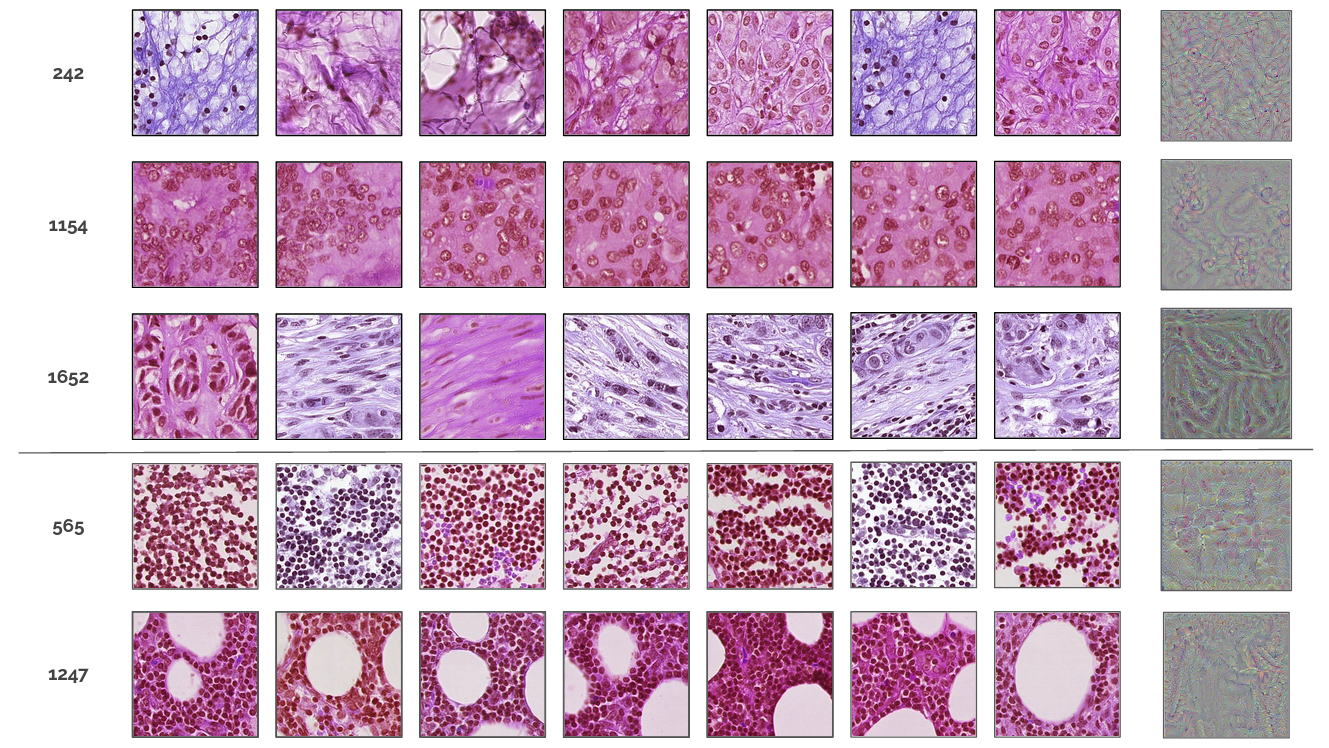}
\caption{Patch-based visualizations obtained for features 242, 1154 and 1652 (for min-scores features); 565 and 1247 (for max-scores features); tiles and max activation images (right).}
\label{patchbasedvisu}
\end{figure}

Figure \ref{patchbasedvisu} shows the 7 tiles that activate the most (over all tiles) each feature and the max activation image, that we expect to reveal what the feature reveals with regards to the histopathological problem it has been trained on. 

% TODO
Pathologists agreed that patch-based tiles visualizations are highly interpretable and reveal features that are indeed related to each class \cite{Tan19}. For example, feature 1652 tends to trigger spindle-shaped cells that indeed can be a metastasic tissue organization. For ``normal" tissue features, feature 565 describes mainly clustered lymphocytes that are preponderant in normal tissues. 

Coherence between patches exposed for a better interpretability led us to think about another way to present features to pathologists. Indeed, since tissues have coherent and somehow organized structure, a relevant feature for histological problems would be activated in a coherent and somehow organized way over slides. Thus, along with patch-based visualization, we propose to access features activation heat-maps $H_{c,i}$ over slides as presented in Section \ref{contrib}.

\begin{figure}[!th]
\centering
\includegraphics[width=0.8\textwidth]{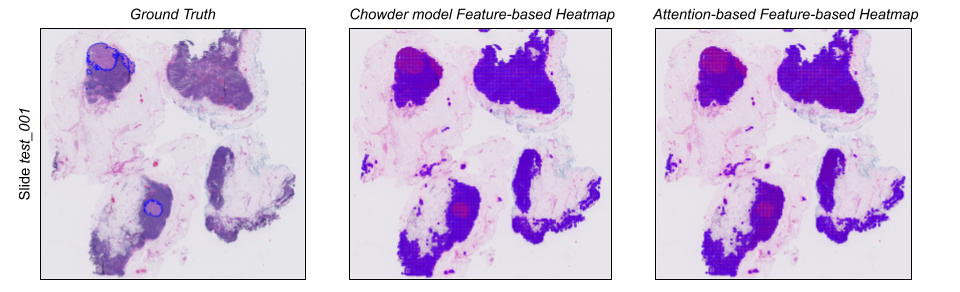}
\caption{Slide-based visualizations: Heat-maps obtained by computing average normalized activation over identified features; ground-truth annotations for ``tumor" tissue (left); CHOWDER model feature-based heat-maps (middle); attention-based model feature-based heat-maps (right).}
\label{slideclasssegmentation}
\end{figure}

Figure \ref{slideclasssegmentation} illustrates qualitative results. Quantitatively, we report an AUC of 0.884 for CHOWDER model and 0.739 for Attention-based model, using this average normalized activation as a ``tumor" predication score and using lesion annotation provided by Camelyon-16 dataset to get the ground-truth label per tile. Both AUCs are significantly high, which validates our approach of identifying features that are relevant and of computing heat-maps for interpretation and explanation. Note that the AUC computed using tile scores is 0.684 for CHOWDER model and 0.421 for Attention-based model (see Table \ref{table_FROC}). We can also note that there is a gap in interpretability between CHOWDER model and Attention-based model while classification performances are comparable. The gap can be explained by the fact that, in the context of Camelyon-16, identifying one tumorous tile is enough to label a slide as ``tumor", so implicit tile classification does not need to be exhaustive to provide meaningful information to the slide level decision module, however if so interpretability will dicrease. 
% \\

\begin{table}[!th]
\caption{Results: classification and localization AUC using both methods (improvement of localization AUC by 29.2\% for CHOWDER and 75.5\% for Attention-based).}
\centering
{\small
\begin{tabular}{|c|c|c|c|}\hline
Model & Classification AUC& Heat-map method & Localization AUC \\ \hline
\multirow{2}{*}{CHOWDER}  & \multirow{2}{*}{0.82} & Tile scores & 0.684\\
&&Feature-based (ours) & {\bf 0.884}  \\ \hline
\multirow{2}{*}{Attention-based}  & \multirow{2}{*}{{\bf 0.83}} & Tile scores & 0.421\\
&&Feature-based (ours) & 0.739 \\ \hline
\end{tabular}
}
\label{table_FROC}
\end{table}

\section{Conclusion}
\label{conclusion}
In this paper, we presented our interpretability approach and researches for WSI classification architectures. We proposed a unified design that gathers a large majority of WSI classification methods relying on MIL learning, and applied a gradient-based attribution method to identify features that have been learned to be relevant in intermediate (tile and slide) descriptors. Then we showed the relevance of these features by visualization, and validated it with the help of pathologists. We finally proposed explainability heat-maps over whole slides taking into account only identified features. This contribution considerably improved tile classification AUC. Allying patch-based and slide-based visualization took interpretability to a next level for pathologists to understand histological meanings of features used by trained models.


\end{document}